%% file: main.tex
\documentclass[journal]{IEEEtran}

\usepackage{cite}
\usepackage{amsmath,amssymb,amsfonts}
\usepackage{algorithmic}
\usepackage{graphicx}
\usepackage{textcomp}
\usepackage{amsmath,bm}
\usepackage{soul} 
\usepackage{multirow}
\usepackage{soul,color,colortbl}
\usepackage{caption}
\usepackage[table,xcdraw]{xcolor}
\usepackage{cite}

\def\BibTeX{{\rm B\kern-.05em{\sc i\kern-.025em b}\kern-.08em
    T\kern-.1667em\lower.7ex\hbox{E}\kern-.125emX}}

\begin{document}

\title{Dataset and Analysis of Long-Term Skill Acquisition in Robot-Assisted Minimally Invasive Surgery}

\author{Yarden Sharon$^{*1,2}$, Alex Geftler$^{*3}$, Hanna Kossowsky Lev$^{1}$, and Ilana Nisky$^{1}$, \textit{Senior Member, IEEE} \vspace{-5mm}
\thanks{This work was supported by the ISF under Grant 327/20, by the BSF under grant 2023022, by the Helmsley Charitable Trust through the Agricultural, Biological and Cognitive Robotics Initiative, by the Marcus Endowment Fund, by SimReC -- the research center for simulation in healthcare, all at Ben-Gurion University of the Negev. Yarden Sharon was supported by the Besor scholarship and the Israeli Planning and Budgeting Committee scholarship.}
\thanks{$*$ Yarden Sharon and Alex Geftler contributed equally.}
\thanks{$^{1}$Department of Biomedical Engineering and Zelman Center for Brain Science Research, Ben-Gurion University of the Negev, Beer-Sheva, Israel.}%
\thanks{$^{2}$Haptic Intelligence Department, Max Planck Institute for Intelligent Systems, Stuttgart, Germany.}%
\thanks{$^{3}$Department of Orthopedic Surgery, Soroka Medical Center, Beer-Sheva, Israel.}       
}

\maketitle

\begin{abstract}
Objective: We aim to investigate long-term robotic surgical skill acquisition among surgical residents and the effects of training intervals and fatigue on performance.
Methods: For six months, surgical residents participated in three training sessions once a month, surrounding a single 26-hour hospital shift. In each shift, they participated in training sessions scheduled before, during, and after the shift. In each training session, they performed three dry-lab training tasks: Ring Tower Transfer, Knot-Tying, and Suturing. We collected a comprehensive dataset, including videos synchronized with kinematic data, activity tracking, and scans of the suturing pads.
Results: We collected a dataset of 972 trials performed by 18 residents of different surgical specializations. Participants demonstrated consistent performance improvement across all tasks. In addition, we found variations in between-shift learning and forgetting across metrics and tasks, and hints for possible effects of fatigue. 
Conclusion: The findings from our first analysis shed light on the long-term learning processes of robotic surgical skills with extended intervals and varying levels of fatigue.
Significance: This study lays the groundwork for future research aimed at optimizing training protocols and enhancing AI applications in surgery, ultimately contributing to improved patient outcomes. The dataset will be made available upon acceptance of our journal submission.
\end{abstract}


\section{Introduction}
In robot-assisted minimally invasive surgery (RAMIS), surgeons manipulate robotic joysticks to control the movements of surgical instruments inside the patient's body via teleoperation \cite{maesoEfficacyVinciSurgical2010}. RAMIS offers many benefits compared to open and laparoscopic surgery \cite{moorthyDexterityEnhancementRobotic2004,lanfrancoRoboticSurgeryCurrent2004,rassweilerLaparoscopicRoboticAssisted2006}. However, the technical skill level of the surgeon remains a critical factor in patient outcomes. Poor surgical skills increase the risk of postoperative complications, prolonged recovery times, and even mortality, highlighting the need for effective surgical training programs \cite{birkmeyerSurgicalSkillComplication2013}. While significant effort is invested in optimizing the acquisition of technical skills for RAMIS surgeons \cite{satavaProvingEffectivenessFundamentals2020, malpaniEffectRealtimeVirtual2020, marianiSkillOrientedPerformanceDrivenAdaptive2021,caccianigaMultiSensoryGuidanceFeedback2021,gomezSimulationTrainingHaptic2024,oquendoHapticGuidanceHaptic2024,choksiStandardizingSurgicalTraining2025}, the existing training protocols have substantial room for improvement. 

Previous surgical training studies have indicated that distributed practice, characterized by brief training sessions spread over multiple days or weeks, is more effective for skill acquisition than massed practice, characterized by prolonged training sessions concentrated within a single day or over two consecutive days \cite{cecilio-fernandesAvoidingSurgicalSkill2018,joostenEffectIntervalTraining2021,fahlBestPracticeDeveloping2023,jorgensenDistributedTrainingVs2025}. However, surgical residents' demanding schedules often prevent consistent training session scheduling, resulting in irregular intervals between practice opportunities \cite{shawCurrentBarriersRobotic2022a}. In hospitals with a single surgical robotic system used for surgeries and training, residents may encounter even fewer opportunities for regular practice. The irregular and infrequent training intervals available to residents raise crucial questions about the nature of their learning: which aspects of skills are retained, which are lost, and which may continue to improve even in the absence of training. Although research has been conducted in this area, most existing studies have focused on short-term intervals between sessions, ranging from days to weeks \cite{jorgensenDistributedTrainingVs2025}. While studies have begun exploring long-term surgical skill acquisition \cite{kumarObjectiveMeasuresLongitudinal2012a,stefanidisMulticenterLongitudinalAssessment2015}, significant gaps remain in understanding these processes in greater detail. Understanding long-term processes could enable the development of better training protocols and virtual reality simulators that accommodate the constraints faced by surgical residents.

In the broader field of motor learning, irregular training schedules similar to those experienced by surgical residents have not been extensively studied. Most motor learning research focuses on consistent practice schedules with training sessions spanning hours to days and primarily investigating simple movements \cite{karniAcquisitionSkilledMotor1998,floyer-leaDistinguishableBrainActivation2005,shmuelofAreWeReady2011,krakauerMotorLearning2019}. This limitation means that the principles derived from motor learning studies may not directly apply to the complex and prolonged training required for mastering surgical skills. Consequently, there is a gap in our understanding of how irregular and long-term training schedules affect the acquisition and retention of motor skills in surgical contexts. While recent studies have expanded the understanding of long-term complex motor skill acquisition, such as video gaming \cite{listmanLongTermMotorLearning2021}, they focus on different types of motor skills rather than the fine manipulation required in surgical contexts, highlighting a need for focused research in this area. Addressing this gap could enhance RAMIS training protocols and contribute valuable insights to the motor learning field to deepen our understanding of fine skill learning over extended periods.

In massed practice, the learning curve typically exhibits rapid performance improvements during the initial training sessions, followed by a gradual slowdown until a plateau is reached \cite{porteVerbalFeedbackExpert2007,mastersImplicitMotorLearning2008,brinkmanVinciSkillsSimulator2013, niskyTeleoperatedOpenNeedle2015,baharSurgeonCenteredAnalysisRobotAssisted2020}. When intervals are introduced between training sessions, we expect to observe several phenomena. (1) Retention, where participants maintain the skills acquired from the last session \cite{censorCommonMechanismsHuman2012}. (2) Between-session forgetting, in which trainees lose some of the skills they acquired during the previous session before the next session occurs \cite{aMotorControlLearning2019}. (3) Between-session (“offline”) learning, where skill improvement occurs during the interval between training sessions without direct practice \cite{meierOfflineConsolidationImplicit2014}. Previous studies have focused on intervals within training programs that spanned days to weeks, leaving the rates of these different processes during extended training durations unclear. Here, we aim to investigate these phenomena over long-term training periods. Potentially, this will provide insights into optimizing protocols of skill acquisition and retention for surgical residents.

When studying surgical interns' long-term skill acquisition, it is essential to consider that their learning often occurs under fatigue conditions. The impact of fatigue on surgical performance has been investigated in various studies \cite{whelehanWouldYouAllow2020,whelehanSystematicReviewSleep2020}; however, significant methodological variations lead to inconclusive findings. Results range from demonstrating a negative effect of fatigue \cite{eastridgeEffectSleepDeprivation2003,gerdesJackBarneyAward2008,yamanyEffectPostcallFatigue2015}, no effect \cite{demariaNightCallDoes2005,veddengImpactNightShifts2014}, to even a positive effect \cite{mickoMicrosurgicalPerformanceSleep2017}. Additionally, motor learning research suggests that sleep influences skill acquisition \cite{walkerPracticeSleepMakes2002,censorCommonMechanismsHuman2012,lemkeCouplingMotorCortex2021}. Therefore, we also aimed to examine the effect of fatigue levels on the long-term processes involved in learning new surgical tasks.

Given the complexity of skill acquisition in such challenging environments, it is critical to leverage available resources to enhance surgical training. Artificial Intelligence (AI) techniques have the potential to significantly advance the fields of surgical robotics and surgical training \cite{maier-heinSurgicalDataScience2017,maier-heinSurgicalDataScience2022,yipArtificialIntelligenceMeets2023,guerreroAdvancingSurgicalEducation2023,levendovicsSurgicalDataScience2024}. However, these fields lack the extensive datasets that have driven progress in other fields. One available resource is the JHU-ISI Gesture and Skill Assessment Working Set (JIGSAWS) \cite{gaoJHUISIGestureSkill2014}, which comprises three dry lab tasks performed by eight surgeons, with 101 recorded trials in total. This dataset is widely used in surgical robotics research for automating skill evaluation and gesture recognition \cite{wangDeepLearningConvolutional2018,xieUnsupervisedApproachMultimodality2024,soleymaniHandsCollaborationEvaluation2024}. However, models trained only on this dataset do not generalize well to clinical-like data \cite{d.itzkovichGeneralizationDeepLearning2022}. A more recent dataset is the Robotics Surgical Maneuvers (ROSMA) dataset \cite{rivas-blancoSurgicalDatasetVinci2023}. It comprises three dry lab tasks performed by 12 participants, with 206 recorded trials in total. Additionally, the BGU Pattern-Cutting Dataset (BGU-PatCut) comprises the pattern-cutting task performed by 54 participants under various conditions, with 1,254 recorded trials in total \cite{sharonAugmentingRobotAssistedPatternCutting2025}. Inspired by these significant datasets, we designed our study to collect and share a new dataset. This dataset will provide additional resources to support the research community in understanding skill acquisition in fine manipulation tasks that are inspired by surgical tasks, as well as advancing AI applications for surgical training and surgical robotics. Expanding the availability of comprehensive and varied datasets will enable the development of more accurate and generalizable AI-driven tools.

In our study, we tracked the learning processes of 18 surgical residents as they acquired robotic surgical skills over six months, with training sessions conducted at approximately one-month intervals. Each month, residents participated in three training sessions surrounding a single 26-hour hospital shift. During each training session, participants performed three surgical dry-lab training tasks. Our contributions are as follows:
\begin{enumerate}

 \item{\textbf{Providing a comprehensive dataset:}} sharing a new dataset to advance research in surgical robotics, surgical skills acquisition, and complex motor learning.
 
    \item {\textbf{Characterizing long-term learning:}}
     quantifying the rates of learning within and between shifts, including retention, between-shifts forgetting, and between-shifts learning in robot-assisted surgical skills with extended intervals between training sessions.
     
      \item {\textbf{Assessing the potential impact of fatigue:}}initial exploration of how fatigue may affect skill acquisition and retention by analyzing training sessions before, during, and after a 26-hour hospital shift.
\end{enumerate}

\section{Methods}
        \subsection{Experimental Setup - the da Vinci Research Kit (dVRK), Training Tasks, and Materials}
\begin{figure} []
				\centering
					\includegraphics[width=\the\columnwidth]{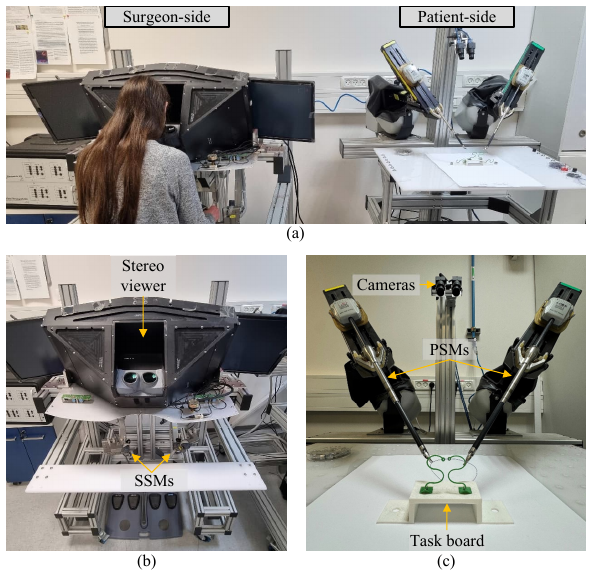}
					\caption[The dVRK]{The da Vinci Research Kit (dVRK). (a) A participant sits on the surgeon side and uses SSMs to control the PSMs on the patient side remotely. (b) Surgeon side configuration. (c) Patient side configuration with the Knot-Tying task.}
					\label{fig:dVRK}
				\end{figure}

        All the experiments in this study were conducted with the da Vinci Research Kit (dVRK). The dVRK is a research platform for RAMIS applications \cite{kazanzidesOpensourceResearchKit2014}, provided by Intuitive Surgical (Sunnyvale, CA) based on the first-generation da Vinci Surgical System \cite{guthartIntuitiveTelesurgerySystem2000}. Our custom-built dVRK setup (Fig. \ref{fig:dVRK}) consists of a pair of surgeon side manipulators (SSMs), a pair of patient side manipulators (PSMs), a foot-pedal tray, a stereo viewer, and four manipulator interface boards. The participants operated the SSMs on the surgeon side (Fig. \ref{fig:dVRK}.b), teleoperating two large needle drivers on the patient side (Fig. \ref{fig:dVRK}.c), where the task boards were placed. The controllers of the SSMs and PSMs were connected via firewire to a single Ubuntu computer equipped with an Intel Xeon E5-2630 v3 processor. The vision system comprised two Blackfly S cameras (FLIR Integrated Imaging Solutions Inc.) to capture the task board in the center of the field of view. The cameras were fixed and not controlled by the participants, and the video was broadcast to the stereo viewer using custom-developed software on a dedicated computer. The movement scale was set to 0.4, allowing participants to reach the entire workspace without using the clutch pedal.
        
\begin{figure*}[]
                         \centering
                         \includegraphics[width=\the\textwidth]{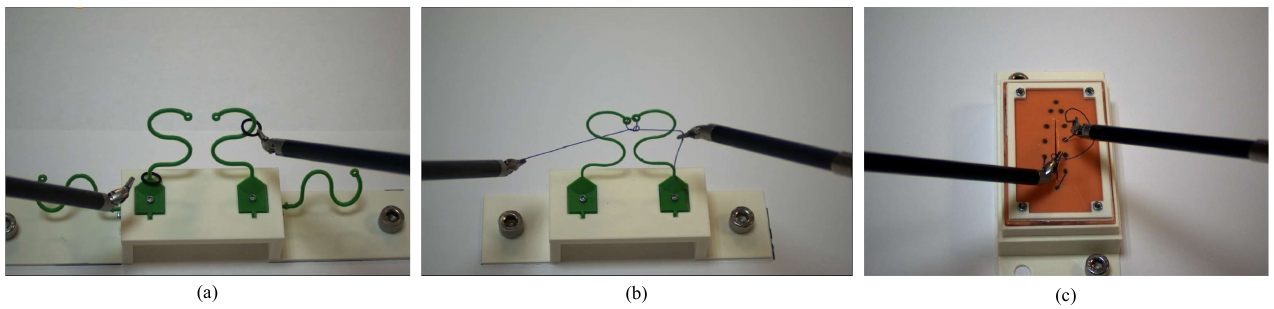}
                         \caption[The surgical training tasks]{The surgical training tasks. (a) Ring Tower Transfer.
                         (b) Knot-Tying.
                         (c) Suturing.}
                         \label{fig:Tasks}
                    \end{figure*}

        In each session, the participants used the dVRK to perform three surgical training tasks based on common tasks for robotic surgeons' training \cite{smithFundamentalsRoboticSurgery2014}:
                               \begin{enumerate}
                  \item \textbf{Ring Tower Transfer} (Fig. \ref{fig:Tasks}.a): the participants transferred the rings from the top towers to the side towers, using each of the right and left tools in turn.
                  \item \textbf{Knot-Tying}  (Fig. \ref{fig:Tasks}.b): the participants tied a surgical knot to attach the two eyelets of the towers.
                  \item \textbf{Suturing} (Fig. \ref{fig:Tasks}.c): the participants sutured a wound in an artificial tissue through marked targets.
                \end{enumerate}

                 Detailed instructions, performance metrics, and critical errors for each task are presented in Videos 1--3.

              We modified the three training tasks to fit the workspace of our dVRK and manufactured all the dry-lab setups that we used in the study in-house. The platforms, towers, and rings were printed using a 3D printer. The suturing tissues were made using two types of silicone: Ecoflex 00-30 Silicone, which simulated a layer of skin, and Ecoflex GEL, which simulated the underlying layer of fat \cite{HowMakeSilicone}. We produced these tissues using 3D-printed molds in two stages: first, preparing the outer layer of `skin' and then filling the inner layer of `fat' using different molds. The shape of the wound was made using the molds. To draw the markers on the tissues, we used a pencil to mark a piece of paper placed on a 3D-printed stamper and then stamped the markers using a template. All the 3D designs and the instructions for preparing the suturing tissues are included in the dataset.
              
              A new tissue was used for each Suturing trial, and a new needle was provided for each resident during each shift, meaning that the same needle was used for three Suturing trials.  For the Knot-Tying and the Suturing task, we used a 2-0 Coated Vicryl Plus CT-2 suture (VCP333H, Ethicon). For the Knot-Tying task, we cut 15 cm sutures. For the Suturing, we tied a ring 15 cm from the needle, designed to prevent the end of the suture from coming out of the start marker.

                \subsection{Participants and Experimental Procedures}
                Twenty-one volunteers (aged 27-40) participated in the study after signing an informed consent form approved by the Human Subjects Research Committee of Ben-Gurion University of the Negev (protocol number 2109-1, approved on July 28, 2021). All the participants were surgical residents from the Soroka Medical Center from different specialties, including cardiothoracic surgery, ENT (ear, nose, and throat) surgery, general surgery, neurosurgery, ophthalmology, orthopedic surgery, pediatric surgery, plastic surgery, urology, and vascular surgery. The residents in the study were at different stages of their residency, ranging from first-year residents to those in their sixth year. None had prior experience with operating RAMIS platforms.

\begin{figure}[]
                         \centering
                         \includegraphics[width=\the\columnwidth]{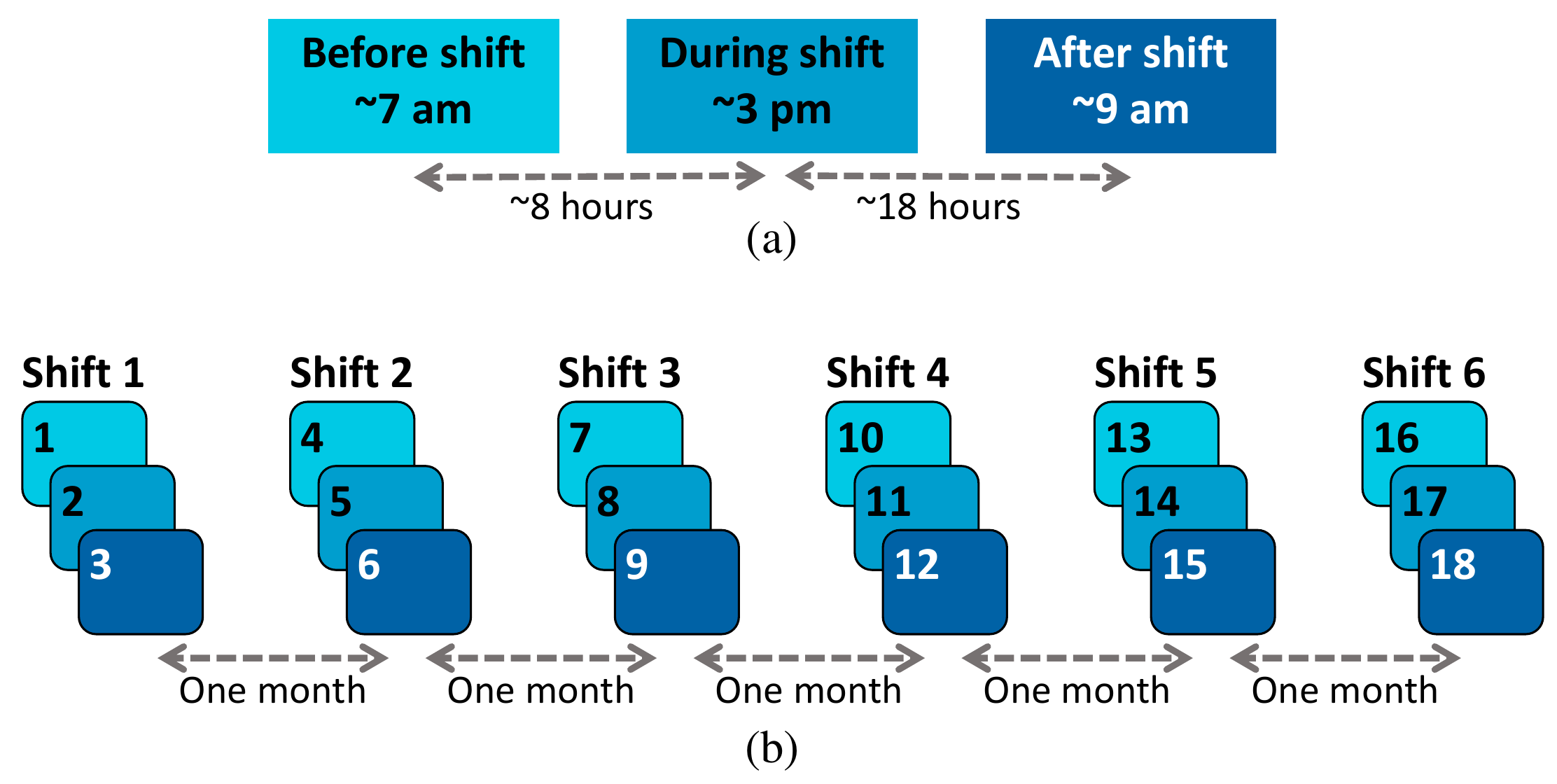}
                         \caption[The protocol of data collection]{Data collection protocol. (a) The three sessions surrounding a single hospital shift. (b) The 18 sessions throughout six months.}
                         \label{fig:CollectionSessions}
                    \end{figure}

               Each resident was scheduled to arrive at the lab once a month over a six-month period. Each month, the resident was intended to participate in three training sessions surrounding a single 26-hour hospital shift (Fig. \ref{fig:CollectionSessions}): before the shift started (approximately 7 am), during the shift (approximately 3 pm on the same day), and after the shift (approximately 9 am the following morning). Nevertheless, the intervals between shifts and the timing within each shift were adjusted throughout the data collection to accommodate the surgical departments' schedules and emergencies in the operating room. 
               
                    A training session lasted up to 30 minutes. It included filling out a questionnaire about activity and fatigue levels, as well as adjusting the SSMs' Velcro straps on the participants' fingers. Participants then proceeded to the training tasks in the same order each time. In the first session of each shift (the morning before the shift), participants also completed a short warm-up task to ensure comfort at the console before proceeding to the training tasks. This warm-up involved using the right tool to move an object from a target on the right side of the workspace to a target on the left side and then retrieving it with the left tool and returning it to the right target.
                    
                    The first training session of the study lasted up to 60 minutes and also included a general explanation of the study and the dVRK system, along with the completion of a demographic and background questionnaire. Participants received videos with the instructions, performance metrics, and critical errors for each task (see Videos 1--3). We confirmed that they watched the videos and understood the instructions. The participants received no summative or formative feedback on their performances during the course of the study.

    \subsection{Data Acquisition, Preprocessing, and Labeling}
\begin{figure}[!t]
         \centering
         \includegraphics[width=\the\columnwidth]{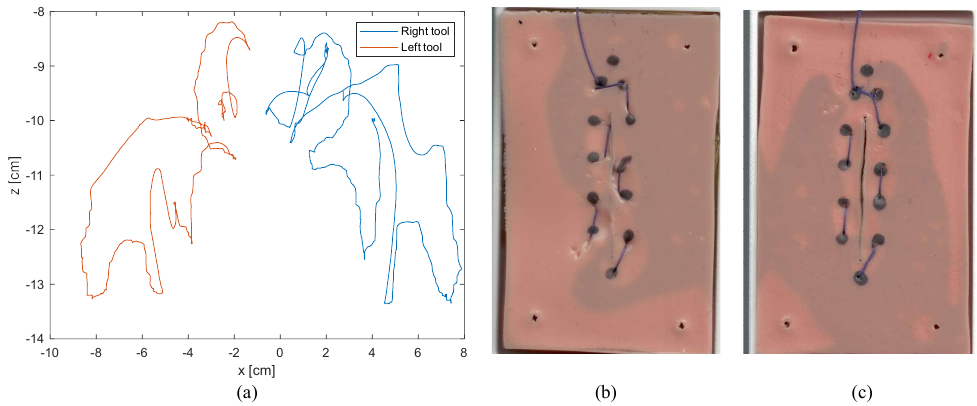}
         \caption[Examples of the recorded data]{Examples of the recorded data. (a) The path of the tools in a Ring Tower Transfer task. (b),(c) Suturing pad scans of two participants.}
         \label{fig:Data}
    \end{figure}

    Using the dVRK, we recorded the kinematic data of the surgical tools at 100 Hz. Fig. \ref{fig:Data}.a shows an example of the tools' path recorded during a Ring Tower Transfer trial. All signals were recorded along with their timestamps, and to synchronize them, we resampled them to identical time points across all the signals at 50 Hz. We resampled the positions, velocities, grippers' opening angles, and wrenches using piecewise cubic Hermite interpolating polynomial (PCHIP). The orientation data were recorded as quaternions and were resampled using spherical linear interpolation (SLERP) \cite{shoemakeAnimatingRotationQuaternion1985}. Before analysis, we filtered the Cartesian PSMs' position data using MATLAB's filtfilt() with a second-order low-pass Butterworth filter and a cutoff frequency of 6Hz. This resulted in zero-phase, fourth-order filtering with a cutoff frequency of 4.81Hz.
    
    In addition, we recorded videos of all the trials from the left camera. There were lost images that were presented in the stereo viewer but not recorded. Therefore, the videos in the dataset were saved using their timestamps in a variable frame rate format.

     For potential future evaluation of the accuracy of the Suturing task, we scanned all the sutured tissues. Fig. \ref{fig:Data}.b-c show examples of tissues scanned during the data collection.

To assess participants' fatigue levels, they completed a questionnaire that included various questions about their activity level, coffee consumption, cigarette use, and other relevant factors. The questionnaire also included the Epworth Sleepiness Scale (ESS) to evaluate their daytime sleepiness \cite{johnsNewMethodMeasuring1991}. They were asked to wear a Xiaomi Mi Smart Band 5 fitness tracker from the day before their shift until the last session after the shift; they did not wear it in the operating room. The questionnaire included sections to report the number of steps and sleep data collected from the fitness tracker app. Although we did not use the activity and fatigue data for the analyses in this paper, the completed questionnaires are included in the dataset for use in further studies.
    
We used the videos, the PSMs' positions, and the grippers' opening angles to manually label the beginning and end of each trial. Additionally, we marked events that occurred during the task and caused breaks in the task sequence, including when the ring fell outside the workspace, the suture came out of the eyelets, got stuck in the tools, or broke. Furthermore, we used the recorded dVRK data to automatically label events where the robot crashed due to teleoperation errors.

    \subsection{Performance Metrics}
     For each trial, we calculated three objective performance metrics to assess surgical skill level:   
    \begin{enumerate}
    \item \textit{Completion time} -- the duration between the start and end of the task. A shorter completion time indicates better performance.
    \item \textit{Path length} -- the total distance traveled by the tips of the PSMs. Since the tasks were bi-manual, we summed the path lengths of both PSMs for each trial to obtain a single metric per trial. A lower path length indicates better performance.
    \item \textit{Rate of orientation change} -- the average rate at which the PSMs' orientations changed, i.e., how quickly the participants rotated their hands. For this metric, we averaged the values from both right and left PSMs. A higher value indicates better performance.
\end{enumerate}

The formulas for calculating these metrics are detailed in our previous work \cite{y.sharonRateOrientationChange2021}. Events that interrupted the performance of the task were removed from the calculations. For example, if the dVRK crashed during a trial, the data from the time of the crash until teleoperation was re-enabled were not included in the metrics calculations. In several trials of the Knot-Tying task, the suture became stuck, necessitating its removal and restarting the trial. In these trials, more than one start point was labeled. For the analyses in this paper, we used the data from the last labeled start point until the end of the trial. 

    \subsection{Statistical Analysis}
    For each task and metric, we used two statistical models to assess the learning processes.
    Model 1 is a two-way repeated measures ANOVA to evaluate the learning across the six shifts and the effect of the timing of the training session relative to the shift on the performance. This model included two within-subject factors: \textbf{Shift Number} and \textbf{Timing Relative to Shift}. \textbf{Shift Number} consisted of six levels, shift 1 to shift 6. \textbf{Timing Relative to Shift} had three levels: before, during, and after the shift -- allowing us to observe a potential influence of fatigue experienced by the residents. Additionally, we examined the interaction between \textbf{Shift Number} and \textbf{Timing Relative to Shift} to determine if the effect of the session timing varied between the shifts.
    
    Model 2 is another two-way repeated measures ANOVA, used to investigate the processes during intervals between the shifts. For this analysis, we used only the last sessions of shifts 1 to 5 and the first sessions of shifts 2 to 6. For example, for the first interval between shifts, we analyzed the last session of shift 1 (before the interval) and the first session of shift 2 (after the interval). This model included two within-subject factors: \textbf{Between Shift Interval Number} and \textbf{Timing Relative to Shift Interval}. \textbf{Between Shift Interval Number} had five levels, intervals 1 to 5, representing the intervals between consecutive shifts. \textbf{Timing Relative to Interval} consisted of two levels: before and after the interval, indicating whether the session occurred before or after the interval between shifts. We also examined the interaction between \textbf{Between Shift Interval Number} and \textbf{Session Timing Relative to Interval} to determine if the between-shift processes varied throughout the study.

    Due to the exploratory nature of this study, we chose not to perform any post hoc or planned comparisons, even when main or interaction effects reached statistical significance. However, the direction of these effects can be inferred from the plots presented or calculated directly from the dataset. These observations may inform hypotheses and guide the design of future, more targeted experiments. 
    
    Participants with any missing values were excluded from the analyses. The numbers of participants included in each task and model are reported in the Results section. Before conducting the ANOVAs, we log-transformed the values of the metrics to correct their non-normal distributions. We assessed the assumption of sphericity using Mauchly's test. When violations were detected, we adjusted the degrees of freedom using Greenhouse-Geisser correction. We performed the statistical analyses using the MATLAB Statistics Toolbox. Statistical signiﬁcance was determined at the 0.05 threshold.

    \section{Results}
\begin{figure}[b]
         \centering
         \includegraphics[width=\the\columnwidth]{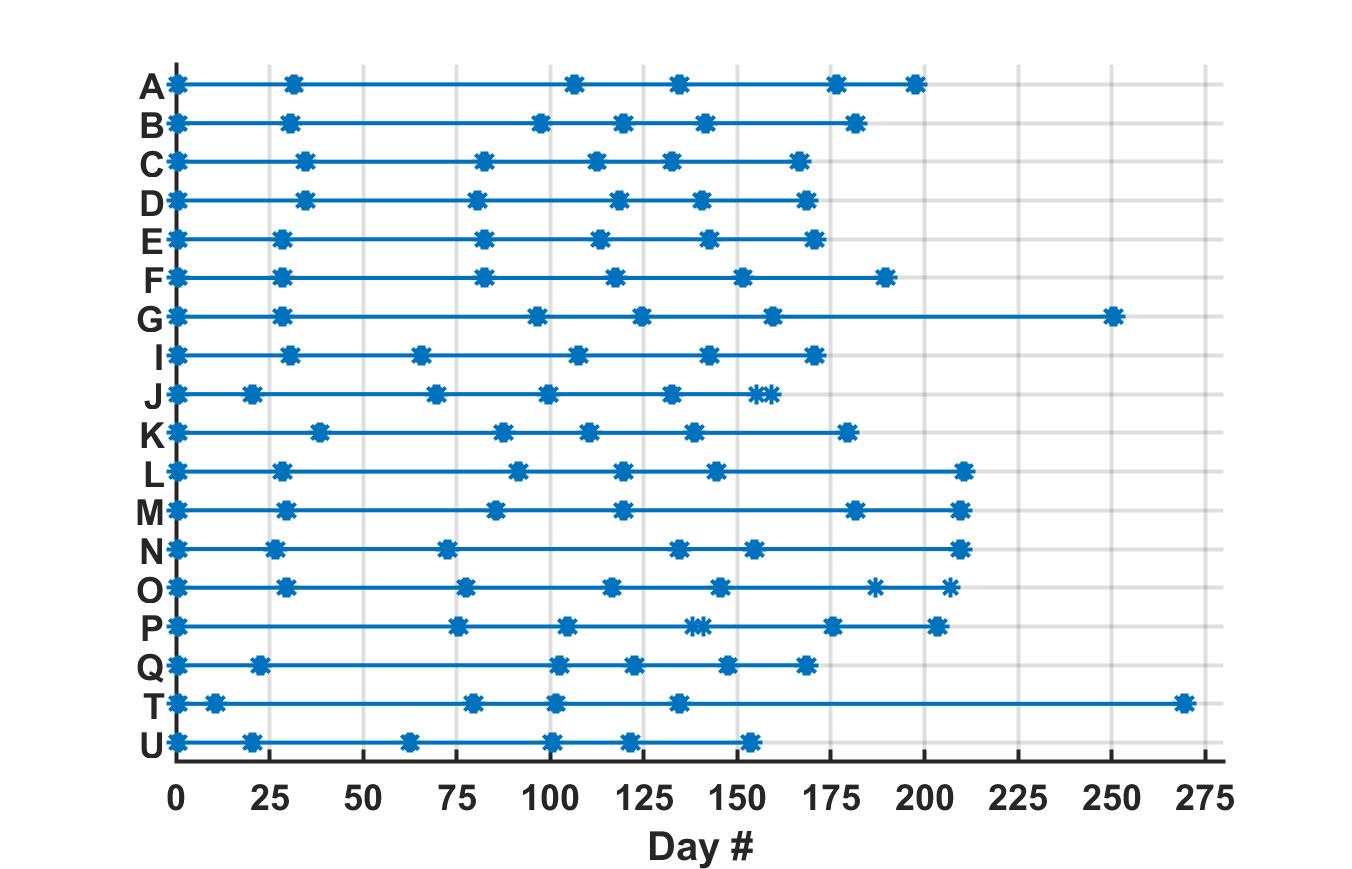}
         \caption{Timeline of training sessions relative to the start date of each participant. Each letter represents a participant who completed the study, and each marker indicates a training session. Note that for sessions conducted in accordance with the protocol (surrounding a single shift), all three markers appear as a single point.}
         \label{fig:shiftscollected}
    \end{figure}

    Eighteen participants completed the entire data collection, each participating in six shifts with three training sessions per shift. This resulted in data from 324 trials per task, totaling 972 trials across the three tasks. Three participants withdrew after one to three shifts. Their 15 training sessions (45 trials) were excluded from the analyses of this study and are not included in the dataset. The training sessions occurred between July 2021 and May 2022. We aimed to have intervals of one month between each session; however, adjustments were necessary due to two COVID-19 waves and scheduling constraints within the departments. Fig. \ref{fig:shiftscollected} presents the timeline of the training sessions for each participant, arranged relative to their first session.
    
        In several cases, participants could not arrive at a training session at a reasonable time. For example, some could not attend the afternoon session on the same day or were unable to arrive at the session after the shift. In such cases, participants attended a training session when they were available again to maintain the total number of trials per shift in accordance with the protocol, as seen, for example, in the last two sessions of participant J in Fig. \ref{fig:shiftscollected}. These trials were included in the dataset but excluded from the analyses for this paper. This resulted in removing eight sessions (24 trials) from the analyses. Moreover, ten trials were excluded from the analyses because participants did not complete the task or did not perform the task according to the instructions: four Ring Tower Transfer trials, five Knot-Tying trials, and one Suturing trial. A list of data collection notes, including trials that were excluded from the analyses, is provided with the dataset.

     \begin{figure*}[h!]
         \centering
         \includegraphics[width=0.95\textwidth]{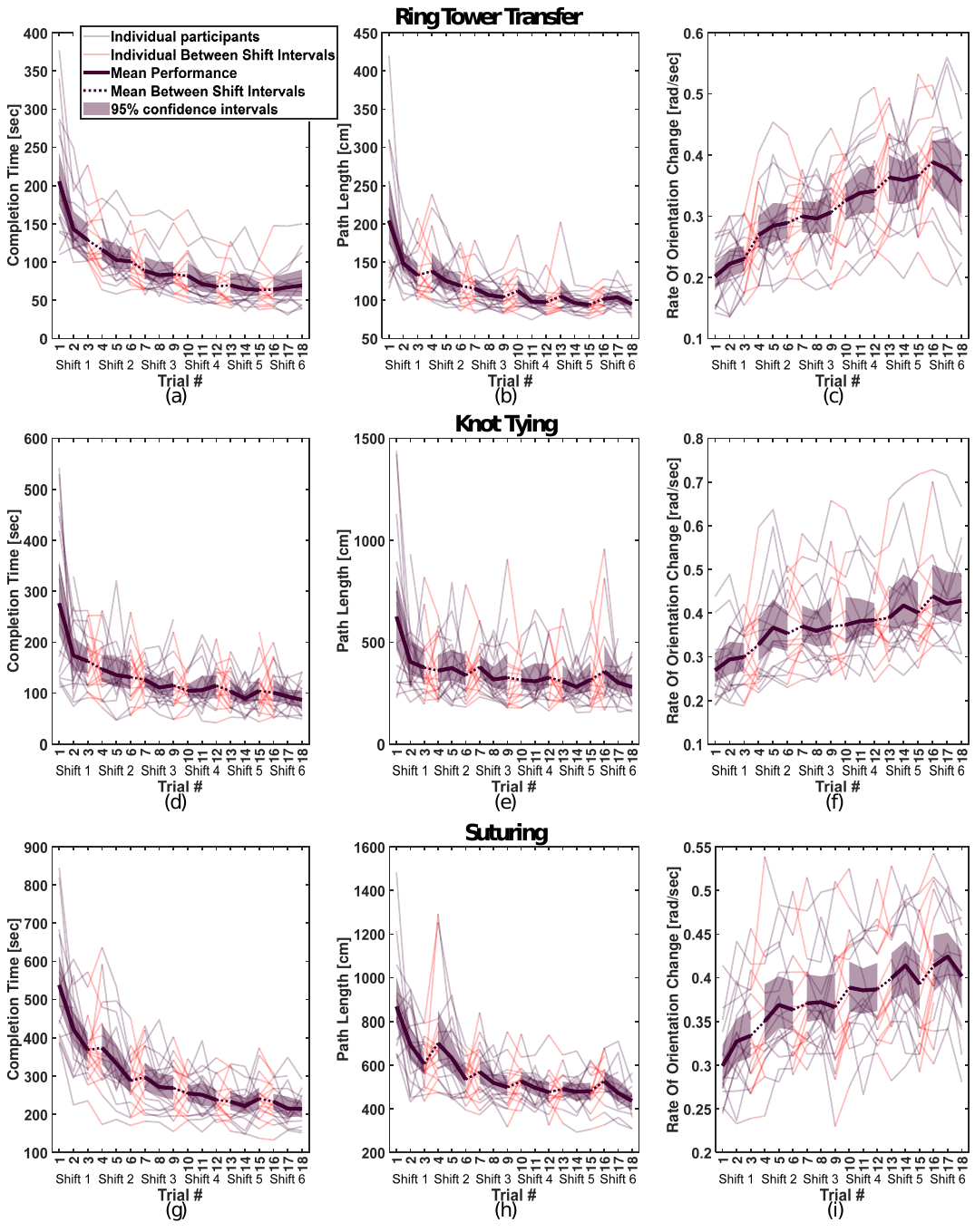}
         \caption{Learning curves throughout the entire data collection sessions for the three tasks (rows) and the three metrics (columns). The plots present the results of all the trials for all the participants: each light purple line represents an individual participant, dark purple lines represent the participants' means, and the shaded areas represent the 95\% bootstrap confidence intervals. The intervals between shifts are marked by red lines for each individual participant and by dashed purple lines for the participants' means.}
         \label{fig:FigAllTrialsResults}
    \end{figure*}
    
\begin{figure}[t]
         \centering
         \includegraphics[width=\the\columnwidth]{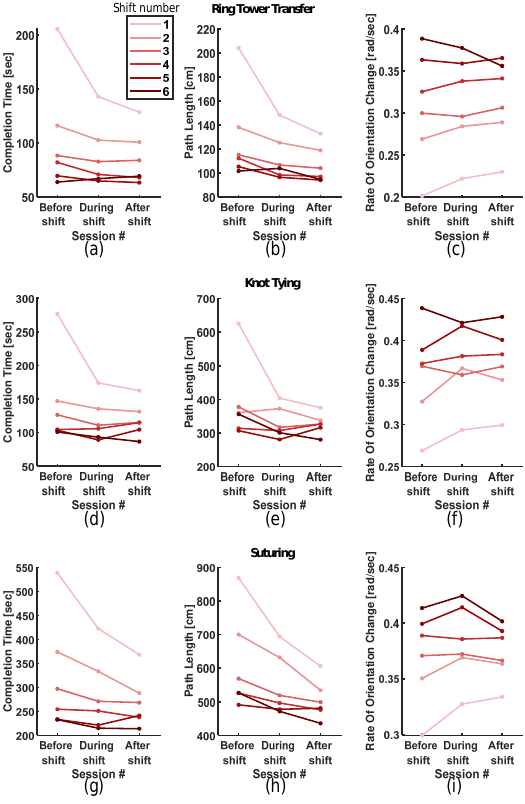}
         \caption{Learning curves throughout the entire data collection sessions for the three tasks (rows) and three metrics (columns). The plots present only the means of the participants, with a distinct curve representing each of the six shifts (1 to 6). The horizontal axis represents the timing within each shift (before the shift, during the shift, and after the shift).}
         \label{fig:ResultsCurvesShifts}
    \end{figure}

\begin{figure}[t]
         \centering
         \includegraphics[width=\the\columnwidth]{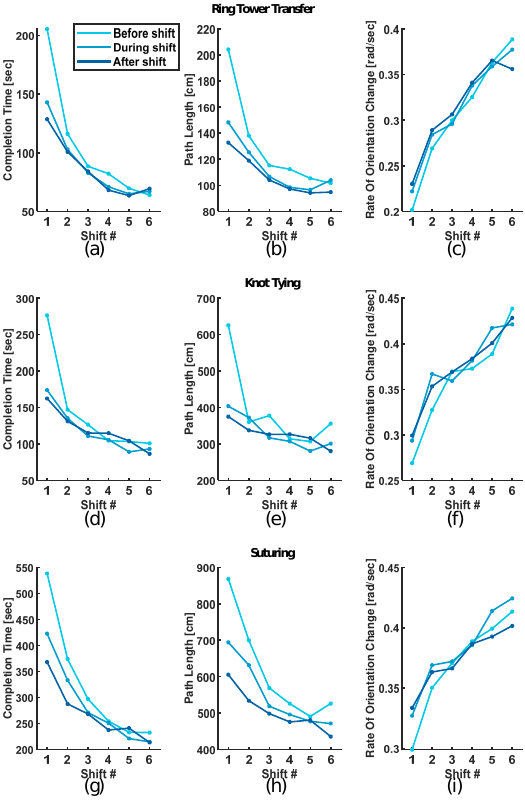}
         \caption{Learning curves throughout the entire data collection sessions for the three tasks (rows) and three metrics (columns). The plots present only the means of the participants, with a distinct curve representing each timing within the shifts (before the shift, during the shift, and after the shift). The horizontal axis represents the shift number (1 to 6).}
         \label{fig:ResultsCurvesSessions}
    \end{figure}

        Figuers \ref{fig:FigAllTrialsResults},\ref{fig:ResultsCurvesShifts}, and \ref{fig:ResultsCurvesSessions} depict the values of the three performance metrics calculated for each task throughout the study, demonstrating that participants improved their performance during the training period. To further analyze these improvements, eighteen two-way repeated measures ANOVAs were conducted: nine models for each primary analysis: 1) overall learning, and 2) between shifts learning. For each analysis, separate models were used for the three tasks---Ring Tower Transfer, Knot-Tying, and Suturing---and each task was evaluated across three performance metrics, resulting in nine models per analysis. Table \ref{table:Statistics} summarizes the results of the statistical analyses.

    \subsection{Model 1: Overall learning}
    The factor of \textbf{Shift Number} was statistically significant across all tasks and metrics, indicating that participants' performances improved from the first to the last shift of the study. 
    The factor of \textbf{Timing Relative to Shift} (before, during, after) was statistically significant for the \textit{completion time} metric in all three tasks, meaning that, on average, participants improved their \textit{completion time} within the shifts. For the \textit{path length} metric, this factor was statistically significant for the Ring Tower Transfer and the Suturing tasks, but not for Knot-Tying. In addition, this factor was not statistically significant for the \textit{rate of orientation change} in all three tasks, indicating that, on average across the shifts, participants did not statistically significantly improve this metric within the shifts.

        The interaction between \textbf{Shift Number} and the \textbf{Timing Relative to Shift} was statistically significant for all metrics in the Ring Tower Transfer and Suturing tasks. This means that the learning within the shifts depended on the shift number. This is demonstrated, for example, in Fig. \ref{fig:ResultsCurvesShifts}.c, where different slopes of the different curves illustrate varying improvement rates between the shifts. In the first shift (palest line), participants showed improvement across the three sessions, whereas in the last shift (darkest line), their performance declined throughout the three sessions. This could be due to the combined effect of a plateau in learning and fatigue; further details are discussed in the Discussion section. 
    
         \begin{table*}[]
     \centering
     	\caption{Statistical Analysis Summary}
         \input{table1}
         \label{table:Statistics}
Results highlighted in gray indicate statistically significant effects (p\textless{}0.05).
\end{table*}
    \subsection{Model 2: Between Shifts Learning}
        For all tasks and metrics, except for the \textit{path length} in the Knot-Tying task, the factor of \textbf{Between Shift Interval Number} was statistically significant. This suggests that performance improved throughout the study, aligning with the overall learning observed in the analyses of Model 1. The disparity in statistical significance of the \textit{path length} in the Knot-Tying task between Models 1 and 2 might be related to the fact that much of the improvement for this metric occurred between the first and second trials, which are not included in Model 2. 
        
The factor of \textbf{Timing Relative to Interval} was included to assess the retention, forgetting, and learning between the shifts and revealed several interesting observations. For the \textit{completion time} metric in the Ring Tower Transfer task and the \textit{rate of orientation change} across all tasks, this factor was statistically significant. This means that participants continued to improve between the shifts and were not negatively impacted by the extended intervals without training. For the \textit{path length} metric, this factor was statistically significant for the Ring Tower Transfer and the Suturing task. As shown in Fig. \ref{fig:FigAllTrialsResults}.b and \ref{fig:FigAllTrialsResults}.h, participants began shifts with worse (higher) \textit{path length} performance than what they achieved at the end of the previous shift, suggesting some degree of forgetting between shifts.

None of the interactions between \textbf{Between Shift Interval Number} and \textbf{Timing Relative to Interval} were significant across all nine models.

\section{Discussion}
This paper presents the collection and first analysis of a dataset aimed at studying long-term robotic surgical skill acquisition. We designed the study to track the performance of surgical residents during three training sessions conducted before, during, and after a single 26-hour hospital shift, with training sessions occurring at one-month intervals over a six-month period. During each session, we collected kinematic data, videos, suturing tissue scans, and activity data.  While approaching the design, we had several potential outcomes in mind, but in general, this is an exploratory study. Specifically, based on existing literature, we did not have detailed hypotheses on how the learning processes would look with three training sessions surrounding one shift under different fatigue conditions and with long intervals between shifts. Hence, in this paper, we focused on providing a detailed account of all the methodological considerations. In addition, we are releasing the collected data, including kinematic data, videos, suturing tissue scans, and activity data. Alongside other publicly available datasets, it has the potential to advance research in surgical robotics, motor learning, and other fields.  

Our first exploratory analysis indicates performance improvement across the six training shifts without clear evidence of a plateau. We also observed variations in between-shift forgetting and learning across different metrics and tasks and the potential effects of fatigue on learning rates.
 These results provide a basis for further studies to better understand the long-term processes of learning surgical skills. In the remainder of this discussion section we propose ideas for future analyses that can be performed on this dataset and future follow-up experiments for detailed hypotheses testing.

Our analyses demonstrated that the surgical residents were able to learn and improve across all three tasks throughout the six training shifts. While this finding may seem expected, it was not guaranteed, given the long breaks between shifts and the fact that part of the training occurred after a 26-hour shift. In addition, we found that between-shifts forgetting and learning processes differ between metrics. For instance, we observed between-shifts learning in the \textit{rate of orientation change}, where participants improved between shifts without additional training, whereas for the \textit{path length}, participants' performance declined between shifts. These differences between the metrics highlight the importance of finding a way to quantify the learning of complex tasks by combining several metrics, as done in simpler tasks by assessing speed and accuracy together \cite{reisNoninvasiveCorticalStimulation2009,niskyEffectsRoboticManipulators2014,m.m.coadTrainingDivergentConvergent2017,oquendoHapticGuidanceHaptic2024}. Moreover, the between-shift patterns also differed across tasks for the same metric. For example, improvement in \textit{completion time} between shifts was significant in the Ring Tower Transfer task but not in Suturing. This highlights that surgical training tasks vary in their characteristics, leading to disparities in how the same performance metric behaves across different tasks. Consequently, to effectively develop a wide range of skills and ensure proper assessment of these skills, surgical training should include diverse tasks.

Interestingly, within each shift, the differences in the \textit{rate of orientation change} were not significant, and significant changes occurred between shifts. As illustrated in Fig. \ref{fig:ResultsCurvesShifts}.i, for example, there were differences in the slopes of learning within most shifts. Specifically, in five of the six shifts, there was improvement from the session before the shift to the session during the shift, followed by either smaller improvements or even a decrease in performance between the session during the shift and the session after the shift. These results may be attributed to the fact that the session after the shift was conducted under conditions of fatigue. This also means that the between-shifts learning we observed for the \textit{rate of orientation change} may be influenced by the fact that the last training session in each shift occurs while participants are fatigued, whereas the first session takes place in the morning before the shift when they are likely to perform at their best. To further test the effect of fatigue, one could check if there is enough data and sufficient variation of sleeping hours we recorded for each session to explore correlations between sleep duration and learning rates. If the collected data is insufficient, it will be necessary to plan a controlled experiment. For example, in an experiment in which some participants arrive after a night shift at the hospital, and others arrive after sleeping at home. It is important to note that in such an experiment, it will be essential to consider the variation between the learning rates among participants. 

When we designed this study, we selected three tasks with varying levels of complexity. Simpler and more structured tasks, such as the Ring Tower Transfer, can be used to effectively test specific hypotheses, while the more complex tasks, such as Suturing and Knot-Tying, provide data closer to clinical situations. This variation in task complexity has implications for performance consistency. Although we did not conduct a formal analysis to confirm this observation, it appears that performance in the Knot-Tying task was less consistent across trials and participants, leading to greater variability in the metrics we analyzed. This greater variability in performance may arise from the nature of the Knot-Tying task, which allows for more freedom of movement and diverse approaches among participants. In contrast, the structured design of the Ring Tower Transfer task limits participants' movements, and the targets in the Suturing task impose some constraints as well, resulting in more consistent performance across those tasks.
Additionally, in both the Ring Tower Transfer and Suturing tasks, participants could continue despite making errors (such as touching the towers in the Ring Tower Transfer or misplacing the needle in Suturing). However, in the Knot-Tying task, participants had to successfully complete a knot before progressing to the next one, adding an additional layer of complexity. When utilizing this dataset for future analyses, it is crucial to select tasks that align with the complexity level required to address specific research questions.

In this paper, we present only the first analysis of the data. We used three basic performance metrics that we calculated on the data from the entire trial. This approach led to greater variability in the metrics, which may mask the learning effects we aimed to observe. Further analyses could explore distinct segments of the movements. For example, the Suturing task can be divided into the four distinct sutures that comprise it.. A more detailed segmentation could involve dividing each movement into smaller segments, such as reaching toward the target or inserting the needle into the tissue. Such segments would enable the tracking of the improvement within a single trial and would allow the calculation of additional metrics that could not be calculated on the entire trial, such as smoothness. We conducted this segmentation; however, due to the size of the dataset, multiple annotators labeled the temporal segments, leading to some inconsistencies between the labels across different trials. We did not test this segmentation as rigorously as the labels for the beginning and end of each task. Consequently, we are not releasing these labels at this time, but they are available upon request from the corresponding author.

Surgical Data Science (SDS) is a developing field with significant potential to enhance the use of surgical robotics technology \cite{vedulaSurgicalDataScience2017,levendovicsSurgicalDataScience2024}. However, the field faces challenges due to the limited availability of relevant datasets \cite{maier-heinSurgicalDataScience2017,maier-heinSurgicalDataScience2022}. To address this gap, we have decided to share the dataset collected in this study to facilitate further research in robotic surgery and the acquisition of robotic technical skills. Along with the collected data, we are also providing the designs of all task platforms and detailed instructions on how to prepare the suturing pads. This will enable researchers to use this dataset to initially test hypotheses and subsequently design their own controlled experiments, potentially incorporating the same tasks we utilized. We believe that the effort in collecting this dataset was worth it and that, together with other datasets \cite{gaoJHUISIGestureSkill2014,rivas-blancoSurgicalDatasetVinci2023,sharonAugmentingRobotAssistedPatternCutting2025}, it will enable research that will advance the many fields, including those of surgical robotics and motor learning. 

\subsection{Limitations and future work}
The participants in this study were surgical residents who performed surgical procedures between the shifts that we tracked. This may have contributed to improvements in their performance in the tasks of our study that we could not control. Additionally, participants had access to the instructional videos. Before the first session of the study, we verified that they had watched the videos and asked at the beginning of each session if they needed a reminder about the instructions. However, we did not require them to watch the instructions again, which could lead to variations in performance between participants who reviewed the instructions before each shift and those who did not.

Another aspect we could not control was the timing of the shifts and sessions. Although we aimed to have intervals of one month between shifts, we had to be very flexible and preferred to have varying gaps rather than risk losing participants. Additionally, the timing of the sessions within each shift varied depending on the residents and specific situations in their departments. For example, the earliest session during the shift started at 11:15 am, while the latest began at 8:05 pm. Furthermore, we could not control the amount of activity and sleep of the participants before and during the shifts. Some residents were able to sleep before the sessions, while others did not sleep at all. The timing of the sessions is documented through task timestamps, and the activity information is included in the questionnaires, both of which are available in the dataset.

The participants did not receive any summative or formative feedback regarding their performances. As a result, they might have optimized certain aspects of the tasks while remaining unaware of others that needed improvement. Feedback is crucial for acquiring correct surgical skills and can significantly affect the results \cite{elboghdadyFeedbackSurgicalEducation2017,malpaniEffectRealtimeVirtual2020,suyinFormativeFeedbackGeneration2021}. However, in this first study, we aimed to solely test the improvement achieved through self-training without additional guidance.

One observation we made is that the participants were generally very competitive and strove to perform their best in the tasks. However, due to the demands of their work, some participants were in a rush to return to the hospital during certain sessions, which could have affected their performance.

Importantly, the metrics we presented in this paper do not quantify the accuracy of the task, for example, the amount of touching the green towers. Because of the speed-accuracy trade-off \cite{fittsInformationCapacityHuman1954}, to get a complete description of participants' improvement, these metrics should be combined with an accuracy metric.

\section{Conclusions}
This paper presents the collection and first analysis of a dataset focused on long-term robotic surgical skill acquisition. Our findings indicate that surgical residents improved their performance across all tasks over six training shifts, despite facing long breaks and varying levels of fatigue. Notably, the learning and forgetting processes differed among metrics and tasks, underscoring the need for combined quantitative approaches to measure surgical skill acquisition effectively. Our results suggest a potential effect of fatigue on the \textit{rate of orientation change} scores. Overall, this dataset and our findings will enable further analyses that advance the fields of surgical robotics, motor learning, and surgical training methodologies.

\section*{Acknowledgement}
We sincerely thank the volunteer surgical residents for their commitment to participating in this study, arriving early in the mornings, juggling training sessions with their departmental commitments during the day, and attending after nights without sleep. Their significant efforts made this research possible. We also thank Tami Matus for her invaluable and dedicated support, often extending beyond regular hours, which was essential to the success of the study. We thank the research assistants Nadav Amitai, Netali Auerbach, Roni Bercovich, Yuval Berechya, Gili Deutsch, Yoav Farawi, Yotam Francis, Sapir Goldring, Yuval Kassif, Alon Lempert, Noa Yamin, and Ehud Zippin for their assistance in collecting and labeling the data.

\input{Refs_bbl.bbl}

\end{document}

%% file: table1.tex
\setlength\tabcolsep{3.5pt} 

\begin{tabular}{cccccccc}
\hline
\multicolumn{2}{|c|}{} &
  \multicolumn{3}{c|}{Model 1: Overall Learning (N =13)} &
  \multicolumn{3}{c|}{Model 2: Between Shifts Learning (N =16)} \\ \cline{3-8} 
\multicolumn{2}{|c|}{\multirow{-2}{*}{Ring Tower Transfer}} &
  \multicolumn{1}{c|}{\begin{tabular}[c]{@{}c@{}}Shift Number\\ (1,2,3,4,5,6)\end{tabular}} &
  \multicolumn{1}{c|}{\begin{tabular}[c]{@{}c@{}}Timing Relative\\ to Shift\\ (Before, During,\\ After the Shift)\end{tabular}} &
  \multicolumn{1}{c|}{Interaction} &
  \multicolumn{1}{c|}{\begin{tabular}[c]{@{}c@{}}Between Shift\\ Interval Number\\ (1,2,3,4,5)\end{tabular}} &
  \multicolumn{1}{c|}{\begin{tabular}[c]{@{}c@{}}Timing Relative\\ to Interval\\ (Before, After\\ the Interval)\end{tabular}} &
  \multicolumn{1}{c|}{Interaction} \\ \hline
\multicolumn{1}{|c|}{} &
  \multicolumn{1}{c|}{F(df1,df2)} &
  \multicolumn{1}{c|}{\cellcolor[HTML]{D0CECE}59.204(2.201,26.41)} &
  \multicolumn{1}{c|}{\cellcolor[HTML]{D0CECE}12.668(2,24)} &
  \multicolumn{1}{c|}{\cellcolor[HTML]{D0CECE}8.217(10,120)} &
  \multicolumn{1}{c|}{\cellcolor[HTML]{D0CECE}51.406(1.899,28.487)} &
  \multicolumn{1}{c|}{\cellcolor[HTML]{D0CECE}5.509(1,15)} &
  \multicolumn{1}{c|}{1.528(4,60)} \\
\multicolumn{1}{|c|}{\multirow{-2}{*}{Completion Time}} &
  \multicolumn{1}{c|}{p} &
  \multicolumn{1}{c|}{\cellcolor[HTML]{D0CECE}\textless{}0.001} &
  \multicolumn{1}{c|}{\cellcolor[HTML]{D0CECE}\textless{}0.001} &
  \multicolumn{1}{c|}{\cellcolor[HTML]{D0CECE}\textless{}0.001} &
  \multicolumn{1}{c|}{\cellcolor[HTML]{D0CECE}\textless{}0.001} &
  \multicolumn{1}{c|}{\cellcolor[HTML]{D0CECE}0.033} &
  \multicolumn{1}{c|}{0.205} \\ \hline
\multicolumn{1}{|c|}{} &
  \multicolumn{1}{c|}{F(df1,df2)} &
  \multicolumn{1}{c|}{\cellcolor[HTML]{D0CECE}38.039(2.959,35.509)} &
  \multicolumn{1}{c|}{\cellcolor[HTML]{D0CECE}38.26(2,24)} &
  \multicolumn{1}{c|}{\cellcolor[HTML]{D0CECE}6.483(10,120)} &
  \multicolumn{1}{c|}{\cellcolor[HTML]{D0CECE}21.539(4,60)} &
  \multicolumn{1}{c|}{\cellcolor[HTML]{D0CECE}6.15(1,15)} &
  \multicolumn{1}{c|}{0.697(4,60)} \\
\multicolumn{1}{|c|}{\multirow{-2}{*}{Path Length}} &
  \multicolumn{1}{c|}{p} &
  \multicolumn{1}{c|}{\cellcolor[HTML]{D0CECE}\textless{}0.001} &
  \multicolumn{1}{c|}{\cellcolor[HTML]{D0CECE}\textless{}0.001} &
  \multicolumn{1}{c|}{\cellcolor[HTML]{D0CECE}\textless{}0.001} &
  \multicolumn{1}{c|}{\cellcolor[HTML]{D0CECE}\textless{}0.001} &
  \multicolumn{1}{c|}{\cellcolor[HTML]{D0CECE}0.025} &
  \multicolumn{1}{c|}{0.597} \\ \hline
\multicolumn{1}{|c|}{} &
  \multicolumn{1}{c|}{F(df1,df2)} &
  \multicolumn{1}{c|}{\cellcolor[HTML]{D0CECE}31.14(2.487,29.84)} &
  \multicolumn{1}{c|}{0.004(2,24)} &
  \multicolumn{1}{c|}{\cellcolor[HTML]{D0CECE}1.926(10,120)} &
  \multicolumn{1}{c|}{\cellcolor[HTML]{D0CECE}28.059(2.216,33.239)} &
  \multicolumn{1}{c|}{\cellcolor[HTML]{D0CECE}10.137(1,15)} &
  \multicolumn{1}{c|}{1.624(4,60)} \\
\multicolumn{1}{|c|}{\multirow{-2}{*}{\begin{tabular}[c]{@{}c@{}}Rate of Orientation\\ Change\end{tabular}}} &
  \multicolumn{1}{c|}{p} &
  \multicolumn{1}{c|}{\cellcolor[HTML]{D0CECE}\textless{}0.001} &
  \multicolumn{1}{c|}{0.996} &
  \multicolumn{1}{c|}{\cellcolor[HTML]{D0CECE}0.048} &
  \multicolumn{1}{c|}{\cellcolor[HTML]{D0CECE}\textless{}0.001} &
  \multicolumn{1}{c|}{\cellcolor[HTML]{D0CECE}0.006} &
  \multicolumn{1}{c|}{0.180} \\ \hline
 &
   &
   &
   &
   &
   &
   &
   \\ \hline
\multicolumn{2}{|c|}{} &
  \multicolumn{3}{c|}{Model 1: Overall Learning (N =10)} &
  \multicolumn{3}{c|}{Model 2: Between Shifts Learning (N =14)} \\ \cline{3-8} 
\multicolumn{2}{|c|}{\multirow{-2}{*}{Knot Tying}} &
  \multicolumn{1}{c|}{\begin{tabular}[c]{@{}c@{}}Shift Number\\ (1,2,3,4,5,6)\end{tabular}} &
  \multicolumn{1}{c|}{\begin{tabular}[c]{@{}c@{}}Timing Relative\\ to Shift\\ (Before, During,\\ After the Shift)\end{tabular}} &
  \multicolumn{1}{c|}{Interaction} &
  \multicolumn{1}{c|}{\begin{tabular}[c]{@{}c@{}}Between Shift\\ Interval Number\\ (1,2,3,4,5)\end{tabular}} &
  \multicolumn{1}{c|}{\begin{tabular}[c]{@{}c@{}}Timing Relative\\ to Interval\\ (Before, After\\ the Interval)\end{tabular}} &
  \multicolumn{1}{c|}{Interaction} \\ \hline
\multicolumn{1}{|c|}{} &
  \multicolumn{1}{c|}{F(df1,df2)} &
  \multicolumn{1}{c|}{\cellcolor[HTML]{D0CECE}10.286(5,45)} &
  \multicolumn{1}{c|}{\cellcolor[HTML]{D0CECE}4.545(2,18)} &
  \multicolumn{1}{c|}{0.827(4.981,44.828)} &
  \multicolumn{1}{c|}{\cellcolor[HTML]{D0CECE}9.536(4,52)} &
  \multicolumn{1}{c|}{0.231(1,13)} &
  \multicolumn{1}{c|}{0.545(4,52)} \\
\multicolumn{1}{|c|}{\multirow{-2}{*}{Completion Time}} &
  \multicolumn{1}{c|}{p} &
  \multicolumn{1}{c|}{\cellcolor[HTML]{D0CECE}\textless{}0.001} &
  \multicolumn{1}{c|}{\cellcolor[HTML]{D0CECE}0.025} &
  \multicolumn{1}{c|}{0.537} &
  \multicolumn{1}{c|}{\cellcolor[HTML]{D0CECE}\textless{}0.001} &
  \multicolumn{1}{c|}{0.639} &
  \multicolumn{1}{c|}{0.703} \\ \hline
\multicolumn{1}{|c|}{} &
  \multicolumn{1}{c|}{F(df1,df2)} &
  \multicolumn{1}{c|}{\cellcolor[HTML]{D0CECE}2.453(5,45)} &
  \multicolumn{1}{c|}{3.944(1.247,11.227)} &
  \multicolumn{1}{c|}{0.756(4.705,42.344)} &
  \multicolumn{1}{c|}{2.418(4,52)} &
  \multicolumn{1}{c|}{1.732(1,13)} &
  \multicolumn{1}{c|}{0.739(4,52)} \\
\multicolumn{1}{|c|}{\multirow{-2}{*}{Path Length}} &
  \multicolumn{1}{c|}{p} &
  \multicolumn{1}{c|}{\cellcolor[HTML]{D0CECE}0.048} &
  \multicolumn{1}{c|}{0.065} &
  \multicolumn{1}{c|}{0.579} &
  \multicolumn{1}{c|}{0.06} &
  \multicolumn{1}{c|}{0.211} &
  \multicolumn{1}{c|}{0.569} \\ \hline
\multicolumn{1}{|c|}{} &
  \multicolumn{1}{c|}{F(df1,df2)} &
  \multicolumn{1}{c|}{\cellcolor[HTML]{D0CECE}26.948(5,45)} &
  \multicolumn{1}{c|}{0.971(1.253,11.276)} &
  \multicolumn{1}{c|}{1.472(5.148,46.332)} &
  \multicolumn{1}{c|}{\cellcolor[HTML]{D0CECE}20.167(4,52)} &
  \multicolumn{1}{c|}{\cellcolor[HTML]{D0CECE}6.121(1,13)} &
  \multicolumn{1}{c|}{0.985(4,52)} \\
\multicolumn{1}{|c|}{\multirow{-2}{*}{\begin{tabular}[c]{@{}c@{}}Rate of Orientation\\ Change\end{tabular}}} &
  \multicolumn{1}{c|}{p} &
  \multicolumn{1}{c|}{\cellcolor[HTML]{D0CECE}\textless{}0.001} &
  \multicolumn{1}{c|}{0.367} &
  \multicolumn{1}{c|}{0.216} &
  \multicolumn{1}{c|}{\cellcolor[HTML]{D0CECE}\textless{}0.001} &
  \multicolumn{1}{c|}{\cellcolor[HTML]{D0CECE}0.028} &
  \multicolumn{1}{c|}{0.424} \\ \hline
 &
   &
   &
   &
   &
   &
   &
   \\ \hline
\multicolumn{2}{|c|}{} &
  \multicolumn{3}{c|}{Model 1: Overall Learning (N =14)} &
  \multicolumn{3}{c|}{Model 2: Between Shifts Learning (N =17)} \\ \cline{3-8} 
\multicolumn{2}{|c|}{\multirow{-2}{*}{Suturing}} &
  \multicolumn{1}{c|}{\begin{tabular}[c]{@{}c@{}}Shift Number\\ (1,2,3,4,5,6)\end{tabular}} &
  \multicolumn{1}{c|}{\begin{tabular}[c]{@{}c@{}}Timing Relative\\ to Shift\\ (Before, During,\\ After the Shift)\end{tabular}} &
  \multicolumn{1}{c|}{Interaction} &
  \multicolumn{1}{c|}{\begin{tabular}[c]{@{}c@{}}Between Shift\\ Interval Number\\ (1,2,3,4,5)\end{tabular}} &
  \multicolumn{1}{c|}{\begin{tabular}[c]{@{}c@{}}Timing Relative\\ to Interval\\ (Before, After\\ the Interval)\end{tabular}} &
  \multicolumn{1}{c|}{Interaction} \\ \hline
\multicolumn{1}{|c|}{} &
  \multicolumn{1}{c|}{F(df1,df2)} &
  \multicolumn{1}{c|}{\cellcolor[HTML]{D0CECE}71.43(5,65)} &
  \multicolumn{1}{c|}{\cellcolor[HTML]{D0CECE}17.699(1.429,18.58)} &
  \multicolumn{1}{c|}{\cellcolor[HTML]{D0CECE}3.703(4.84,62.914)} &
  \multicolumn{1}{c|}{\cellcolor[HTML]{D0CECE}33.395(4,64)} &
  \multicolumn{1}{c|}{0.995(1,16)} &
  \multicolumn{1}{c|}{0.613(4,64)} \\
\multicolumn{1}{|c|}{\multirow{-2}{*}{Completion Time}} &
  \multicolumn{1}{c|}{p} &
  \multicolumn{1}{c|}{\cellcolor[HTML]{D0CECE}\textless{}0.001} &
  \multicolumn{1}{c|}{\cellcolor[HTML]{D0CECE}\textless{}0.001} &
  \multicolumn{1}{c|}{\cellcolor[HTML]{D0CECE}0.006} &
  \multicolumn{1}{c|}{\cellcolor[HTML]{D0CECE}\textless{}0.001} &
  \multicolumn{1}{c|}{0.333} &
  \multicolumn{1}{c|}{0.655} \\ \hline
\multicolumn{1}{|c|}{} &
  \multicolumn{1}{c|}{F(df1,df2)} &
  \multicolumn{1}{c|}{\cellcolor[HTML]{D0CECE}32.436(5,65)} &
  \multicolumn{1}{c|}{\cellcolor[HTML]{D0CECE}35.417(2,26)} &
  \multicolumn{1}{c|}{\cellcolor[HTML]{D0CECE}2.242(10,130)} &
  \multicolumn{1}{c|}{\cellcolor[HTML]{D0CECE}18.693(4,64)} &
  \multicolumn{1}{c|}{\cellcolor[HTML]{D0CECE}12.828(1,16)} &
  \multicolumn{1}{c|}{0.463(4,64)} \\
\multicolumn{1}{|c|}{\multirow{-2}{*}{Path Length}} &
  \multicolumn{1}{c|}{p} &
  \multicolumn{1}{c|}{\cellcolor[HTML]{D0CECE}\textless{}0.001} &
  \multicolumn{1}{c|}{\cellcolor[HTML]{D0CECE}\textless{}0.001} &
  \multicolumn{1}{c|}{\cellcolor[HTML]{D0CECE}0.019} &
  \multicolumn{1}{c|}{\cellcolor[HTML]{D0CECE}\textless{}0.001} &
  \multicolumn{1}{c|}{\cellcolor[HTML]{D0CECE}0.002} &
  \multicolumn{1}{c|}{0.763} \\ \hline
\multicolumn{1}{|c|}{} &
  \multicolumn{1}{c|}{F(df1,df2)} &
  \multicolumn{1}{c|}{\cellcolor[HTML]{D0CECE}33.436(2.902,37.732)} &
  \multicolumn{1}{c|}{2.038(2,26)} &
  \multicolumn{1}{c|}{\cellcolor[HTML]{D0CECE}2.444(10,130)} &
  \multicolumn{1}{c|}{\cellcolor[HTML]{D0CECE}13.277(4,64)} &
  \multicolumn{1}{c|}{\cellcolor[HTML]{D0CECE}11.63(1,16)} &
  \multicolumn{1}{c|}{1.18(4,64)} \\
\multicolumn{1}{|c|}{\multirow{-2}{*}{\begin{tabular}[c]{@{}c@{}}Rate of Orientation\\ Change\end{tabular}}} &
  \multicolumn{1}{c|}{p} &
  \multicolumn{1}{c|}{\cellcolor[HTML]{D0CECE}\textless{}0.001} &
  \multicolumn{1}{c|}{0.151} &
  \multicolumn{1}{c|}{\cellcolor[HTML]{D0CECE}0.01} &
  \multicolumn{1}{c|}{\cellcolor[HTML]{D0CECE}\textless{}0.001} &
  \multicolumn{1}{c|}{\cellcolor[HTML]{D0CECE}0.004} &
  \multicolumn{1}{c|}{0.328} \\ \hline
\end{tabular}